\documentclass[letterpaper, 10 pt, conference]{ieeeconf}  % Comment this line out if you need a4paper

\IEEEoverridecommandlockouts                              % This command is only needed if 
\usepackage{graphics} % for pdf, bitmapped graphics files
\usepackage{epsfig} % for postscript graphics files
\usepackage{mathptmx} % assumes new font selection scheme installed
\usepackage{times} % assumes new font selection scheme installed
\usepackage{amsmath} % assumes amsmath package installed
\usepackage{amssymb}  % assumes amsmath package installed
\usepackage{subcaption}
\usepackage{float}
\usepackage{url} 
\usepackage{multirow}
\usepackage{fancyhdr}

\title{\LARGE \bf
Visual-Lidar Map Alignment for Infrastructure Inspections
}

\author{Jake McLaughlin$^{1}$ , Nicholas Charron$^{2}$ and Sriram Narasimhan$^{3}$% <-this % stops a space
\thanks{$^{1}$Jake McLaughlin is with Department of Mechanical and Mechatronics Engineering,
        University of Waterloo, Waterloo, Ontario, Canada.
        {\tt\small jfmclaug@uwaterloo.ca}}%
\thanks{$^{2}$Nicholas Charron is with Department of Mechanical and Mechatronics Engineering,
        University of Waterloo, Waterloo, Ontario, Canada.
        {\tt\small  nicholas.c.charron@gmail.com}}%
\thanks{$^{3}$Sriram Narasimhan with the Departments of Civil \& Environmental Engineering and Mechanical \& Aerospace Engineering, University of California, Los Angeles,
        California, USA.
        {\tt\small snarasim@ucla.edu}}%
}

\fancypagestyle{FirstPage}{
\lfoot{This work has been submitted to the IEEE for possible publication. Copyright may be transferred without notice, after which this version may no longer be accessible.} 
}

\begin{document}

\maketitle
\thispagestyle{empty}
\pagestyle{empty}

%%%%%%%%%%%%%%%%%%%%%%%%%%%%%%%%%%%%%%%%%%%%%%%%%%%%%%%%%%%%%%%%%%%%%%%%%%%%%%%%
\begin{abstract}

Routine and repetitive infrastructure inspections present safety, efficiency, and consistency challenges as they are performed manually, often in challenging or hazardous environments. They can also introduce subjectivity and errors into the process, resulting in undesirable outcomes. Simultaneous localization and mapping (SLAM) presents an opportunity to generate high-quality 3D maps that can be used to extract accurate and objective inspection data. Yet, many SLAM algorithms are limited in their ability to align 3D maps from repeated inspections in GPS-denied settings automatically. This limitation hinders practical long-term asset health assessments by requiring tedious manual alignment for data association across scans from previous inspections. This paper introduces a versatile map alignment algorithm leveraging both visual and lidar data for improved place recognition robustness and presents an infrastructure-focused dataset tailored for consecutive inspections. By detaching map alignment from SLAM, our approach enhances infrastructure inspection pipelines, supports monitoring asset degradation over time, and invigorates SLAM research by permitting exploration beyond existing multi-session SLAM algorithms.

\end{abstract}

\thispagestyle{FirstPage}
%%%%%%%%%%%%%%%%%%%%%%%%%%%%%%%%%%%%%%%%%%%%%%%%%%%%%%%%%%%%%%%%%%%%%%%%%%%%%%%%%
\section{INTRODUCTION}
\label{section:intro}

Infrastructure maintenance is vital for modern societies, ensuring the safety of various critical infrastructures like roads, bridges, and power grids. To properly maintain such infrastructure, periodic inspections of their condition and tracking of defects and other vulnerabilities over time are vital. The common practice for such inspections is through visual means, where personnel carry out inspections. Where warranted, such inspections are supported through defect measurements, often carried out manually. These inspections tend to produce quantitative estimates of infrastructure health and frequently result in sub-optimal outcomes. 

Robotics and automation offer a promising solution for enhancing traditional manual inspection methods. Automated inspections improve precision and accuracy in infrastructure maintenance by leveraging 3D mapping technology, which creates digital replicas of physical assets \cite{sdic_bridge_inspection, uav_reconstruction, sdic_ai_defect_quantification}. Automated inspection processes typically involve three steps: data collection, 3D mapping, and inspection. Firstly, various sensors, including cameras and lidar scanners, gather detailed information about infrastructure assets. This data is then used to generate a 3D model of the asset using 3D mapping algorithms, which typically fall into 3 main categories: visual-based \cite{orb-slam3, vins-mono}, lidar-based \cite{lio-sam, loam} or a combination of both \cite{lvi-sam}, each with their advantages and disadvantages depending on the environment and scenario. Lastly, various automated inspection operations can be performed using the 3D model and the visual information, such as segmenting structural components and detecting and measuring defects like spalls. These aspects have been studied in earlier works, e.g., \cite{sdic_ai_defect_quantification}.

Accurate assessment requires aligning data gathered over repeated inspections that span several years, which is often challenging without a global reference frame such as GPS. Existing methods of aligning maps across time without GPS involve localization in a prior map \cite{hloc, quatro, segmatch}, or multi-session SLAM (MSS) \cite{orb-slam3, maplab2, rtabmap, ltmapper, cartographer}. Localization in a prior map aims to determine the position of new sensor information with respect to a given map, whereas multi-session SLAM aims to generate a single unified map that is updated over time as the environment is revisited. Localization is limited by its inability to perform well under significant environmental changes, since it treats the given map as static. MSS addresses this by treating each mapping session as independent, but each contributing to a single consistent map. However, MSS is limited by the dependence on its own front-end SLAM algorithm, limiting its use with other SLAM methods. This constraint is typically in place so the data is represented in a unified manner and can be used across sessions. While this can provide benefits in terms of accuracy, it hurts the generalizability of their methods, and requires all raw data to be kept and reprocessed which may be challenging as data formats change over the years between inspections. Additionally, existing localization and MSS methods employ visual or lidar data for place recognition but do not utilize them together. Using a single sensor modality is often insufficient for the varied environments seen during infrastructure inspections. For example, vision typically struggles in texture-less (e.g., large concrete infrastructure) and poorly lit (e.g., bridge soffit inspection) environments, whereas lidar struggles in highly reflective scenarios (e.g., bridges over water) and in dense foliage (e.g., bridge undersides). 

This study addresses these challenges by proposing an algorithm tailored for repeated infrastructure inspection tasks, leveraging visual and lidar relocalization methods for offline 3D map alignment. A robust and accurate way to align 3D maps generated from mobile mapping systems allows independent inspections to be combined into a consistent time and spatial reference frame, enabling inspectors to track infrastructure deterioration rates accurately. The main contributions of this work are:

\begin{itemize}
    \item A SLAM-agnostic map alignment algorithm that utilizes visual and lidar data to maximize its robustness to different environments.
    \item A non-rigid map alignment formulation that considers varying degrees of drift between maps, maximizing local map overlap, tailored specifically for long-term infrastructure inspection monitoring.
\end{itemize}

\begin{figure*}[!h]
    \centering
    \vspace{7pt}
    \includegraphics[width=0.95\textwidth]{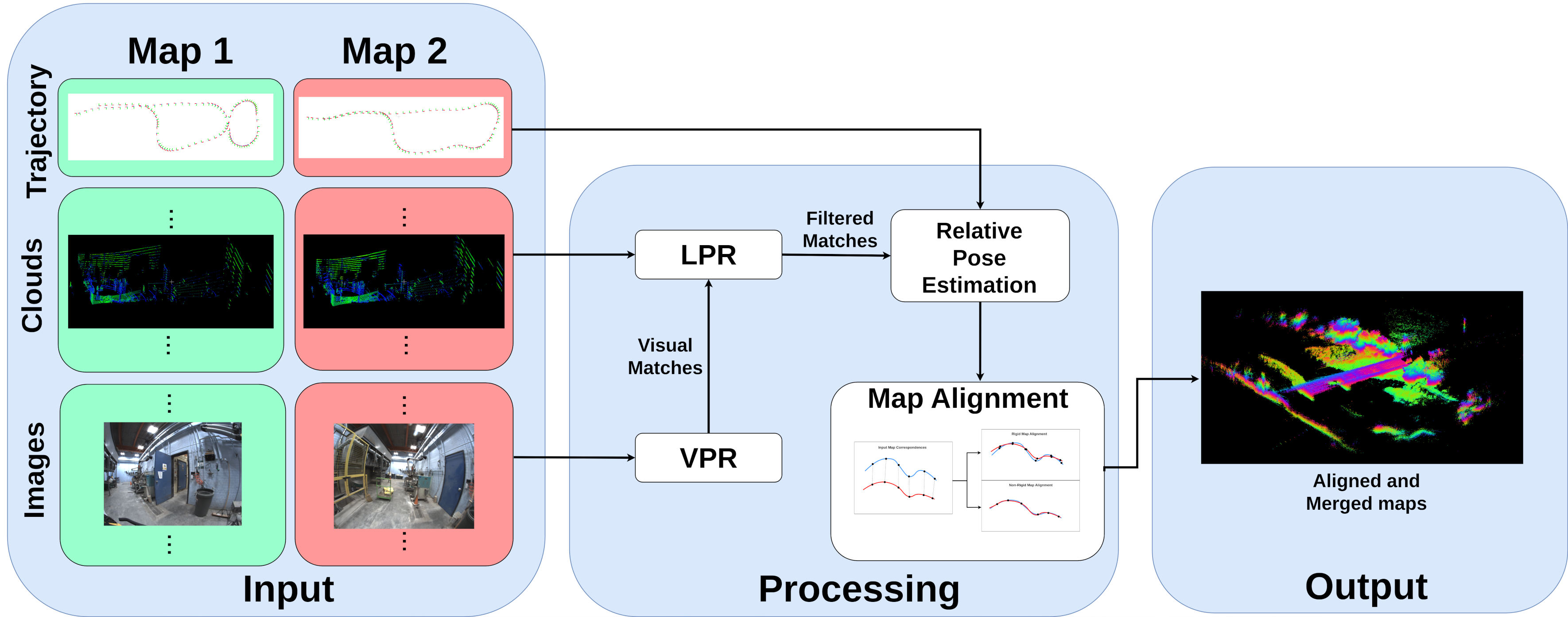}
    \caption{Map alignment pipeline starting from input (a set of images, point clouds and trajectory), to visual place recognition (VPR) and lidar place recognition (LPR) modules which propose matches between maps, to relative pose estimate and finally map alignment to produce the output (a merged point cloud map).}
    \label{fig:mapalignment}
\end{figure*}

\section{RELATED WORKS}
\label{section:related-works}
3D mapping algorithms have been utilized to a great extent in automating infrastructure inspections. In \cite{sdic_bridge_inspection}, an unmanned ground vehicle utilizes various sensors, including GPS, IMU, and LIDAR systems, for creating a detailed 3D map. The study \cite{sdic_ai_defect_quantification} further builds upon this, integrating neural networks for detecting and quantifying structural defects. A popular approach for 3D mapping in the infrastructure space is using drones along with aligned structure-from-motion (SFM) \cite{uav_reconstruction}. However, relatively little work has been done in the space with regard to developing robust means of aligning 3D maps over time for long-term infrastructure assessments.

Although methods exist for aligning 3D maps over time, they have several downsides. The first method relies on a previous map to perform localization using new data. \cite{hloc} performs global localization of images by combining deep place recognition and direct feature matching against structure-from-motion (sfm) map. Using lidar, \cite{quatro} introduces a global registration method to globally localize a 3D scan in a prior lidar map, although it is limited to urban driving scenarios. Localization typically assumes a static map and does not update the map itself during operation. It's useful for precise localization within a known environment but does not account for varying levels of drift in the prior map, nor does it handle changes in the environment over time. Furthermore, pure localization methods relying on a single modality, such as vision or lidar, face limitations in environments where the chosen modality is insufficient. For example, vision-based methods struggle in low-light conditions or environments with sparse visual features, while lidar-based methods are compromised in the presence of highly reflective surfaces or occlusions. Additionally, localization methods are typically designed with specific environments in mind, such as urban driving or indoor spaces. \cite{global_lidar_loc} provides a review of existing lidar-based global localization methods, concluding that existing methods lack generalizability to diverse environments. Similarly, visual localization is trained on static visual datasets, making it very good in the environments it is trained in, but do not generalize to new environments.

An alternate approach is through multi-session SLAM. MSS extends traditional SLAM algorithms to handle repeated visits to the same area over time. Instead of treating each visit as independent, MMS aims to align data from different sessions into a consistent map. This involves merging information from various sessions while accounting for changes in the environment, such as construction or degradation of infrastructure. Most of the existing MSS solutions available today are vision-centric. \cite{rtabmap} is a graph-based visual slam system capable of performing multi-session mapping, through appearance based relocalization techniques. ORB-SLAM3 \cite{orb-slam3}, a 2021 evolution of ORB-SLAM \cite{orbslam} and ORB-SLAM2 \cite{orbslam2}, incorporates IMU integration and multi-session capability. Like \cite{rtabmap}, it uses visual cues for loop closures. One distinction is its map-merging feature. Visual methods like ORB-SLAM3 may face challenges in repetitive or dynamic areas, the former often found in infrastructure applications where structural designs across a system are repeated. Maplab \cite{maplab}, arguably the most comprehensive multi-session mapper, was launched in 2018 and updated in 2022 \cite{maplab2}. It separates odometry and mapping, letting users pick their preferred odometry. While it shares some limitations with RTAB-Map in dynamic environments, its design promotes easy integration of any place recognition algorithm. Comparatively, open-source LiDAR SLAM lacks multi-session capabilities, especially for 3D LiDAR. Cartographer \cite{cartographer} provides both 2D and 3D LiDAR SLAM but limits multi-session features to 2D LiDAR. LT-Mapper \cite{ltmapper}, a comprehensive 3D LiDAR SLAM solution, uses the Scan Context \cite{scancontext} descriptor for place recognition without integrating visual information. Multi-session mapping approaches often come with built-in SLAM algorithms tailored to their specific framework. While these built-in SLAM algorithms may work well within the chosen framework, they are incompatible with other SLAM approaches or algorithms. This limits the flexibility of users in terms of the choice of SLAM algorithms suited to their specific needs or sensor configurations. Additionally, there exists no MSS algorithm that fuses both visual and lidar data, which is essential for accurate and robust mapping in infrastructure environments.
 
Both localization and MSS utilize place recognition to identify revisited locations using sensory data. There are two main categories of place recognition algorithms: knowledge-based and learning-based. One notable knowledge-based visual method is the Dynamic Bag of Words (DBoW) \cite{dbow2}. In DBoW, images are treated as collections of "visual words" made from specific feature descriptors. These words form a vocabulary, allowing new images to be represented as histograms of these words. DBoW is dynamic, adapting the vocabulary as new images/features appear, making it useful for real-time applications, such as autonomous driving. On the deep learning side, NetVLAD \cite{netvlad} aggregates local image descriptors into a compact Vector of Locally Aggregated Descriptors (VLAD), a fixed-size global descriptor. Integrated into a trainable Convolutional Neural Network (CNN), the entire network is optimized end-to-end, enhancing place recognition performance. Scan Context \cite{scancontext} is a prevalent knowledge-based algorithm that transforms 3D LiDAR data into a 2D matrix descriptor. The descriptor offers a panoramic view of the environment, facilitating efficient place recognition. In contrast, BoW3D \cite{bow3d} provides bag-of-words functionality for LiDAR data, and PointNetVLAD \cite{pointnetvlad} is a deep learning method to generate global descriptors of 3D point clouds for large-scale environment recognition. Visual place recognition often outperforms lidar place recognition in certain scenarios, and vice versa. Yet, fusing both modalities of place recognition remains an under-explored area of research.

\section{METHODOLOGY}
\label{section:meth}

The proposed map alignment algorithm consists of four key steps: 1) visual place recognition (VPR) candidate search, 2) lidar place recognition (LPR) outlier rejection, 3) map-to-map scan registration, 4) non-rigid trajectory alignment. This section will outline the details of each step. The complete pipeline for the algorithm is illustrated in figure \ref{fig:mapalignment}. The pipeline is modular, allowing each step to be replaced with a different method (e.g., replacing VPR, LPR and scan registration methods). Map inputs contain 3 sets of data: images, point clouds, and a trajectory, each can be queried by timestamp. The following conventions will be used to refer to the data in each map:

\begin{itemize}
    \item $M_n$ : Map $n$
    \item $I_{M_n}^i$: Image from $M_n$ at timestamp $i$
    \item $S_{M_n}^i$: Point cloud from $M_n$ at timestamp $i$
    \item $T_{L_i}^{W_n}$: Transformation matrix from lidar frame at timestamp $i$ to $M_n$'s world frame
    \item $[t_n^0, t_n^1, \dots, t_n^m]$: Set of timestamps from $M_n$
\end{itemize}

\subsection{Visual Place Recognition}

The VPR candidate search aims to perform an initial search between the images in each map and propose a list of candidate matches. The candidate matches take the form of a list of timestamp pairs, $(t_1^{i}, t_2^{j}$), which are associated timestamps from the reference map $M_R$ and target map $M_T$. The main structure of VPR is to build a visual database for map $M_i : i=1$, which can then be queried by the images in subsequent maps $M_j: j= 1, ..., N$. Instead of adding all images in the reference map to the image database, only spatially separated images are used, which reduces the number of redundant images in the database. To achieve this, we utilize the input trajectory of the reference map and only add images to the database if they have moved over 2 meters or rotated over 45$^{\circ}$ since the last image that has been added. This ensures that there has been a significant visual change since the last image was added to the database and reduces image redundancy.

\begin{figure}[!h]
    \centering
    \includegraphics[width=0.45\textwidth]{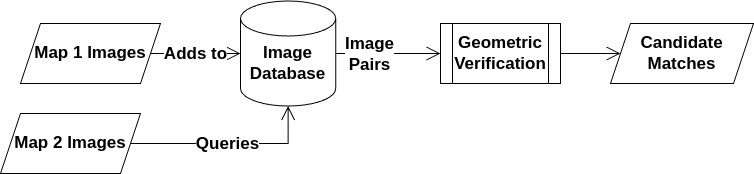}
    \caption{Visual place recognition flow chart, starting from the sets of images in each map (left), building and querying the database, and finally verifying matches.}
    \label{fig:vpr}
\end{figure}

To store and query images, the method proposed in \cite{dbow2} is used. This method involves transforming each image into a histogram of visual words. The process includes computing ORB \cite{orb} descriptors for each detected feature point and assigning each descriptor to a word in the visual vocabulary, which is then accumulated in its histogram of words and added to the database. To make the system more robust, rather than using a pre-trained vocabulary such as the one used in \cite{dbow-reloc}, which is trained on the RAWSEEDS \cite{rawseeds} dataset, we directly train a custom vocabulary on the features extracted from the images in the reference map.

Once the image database is created for the reference map, it is queried with each image from the target map. Each query will return the image from the database that has the most similar histogram of words, as well as the computed similarity score, $S$. If this score is greater than a preset threshold, $\alpha$, it is considered for the next step, which is geometric verification.

Geometric verification is the process of identifying whether two given images are looking at the same place. This is done by first extracting visual features using ORB \cite{orb} detection and description, and then using a brute-force matching method. Once features matches have been extracted between images the essential matrix \cite{multiviewgeometry} between the images is estimated using a combination of Random Sample and Consensus (RANSAC) and the 7-point algorithm \cite{7point}, which is decomposed into a rotation and relative translation (not to scale) using singular value decomposition. Using this relative transformation, we compute the total number of inlier features (their triangulated position projects near its detected pixel in each image). If the percentage of inliers (out of all features matched between images) is over a threshold $\phi$, then geometric verification has passed. If an image pair passes geometric verification it will be inserted into the candidate match list (equation \ref {eq:4.3}) as a timestamp pair, $(t_1^{i}, t_2^{j}$).

\begin{equation} \label{eq:4.3}
C_{M_R}^{M_T} = \left[ (t_1^{i1}, t_2^{j1}), (t_1^{i2}, t_2^{j2}), \dots, (t_1^{in}, t_2^{jn}) \right]
\end{equation}

\subsection{Lidar Place Recognition} \label{sec:lpr}

The LPR module only takes the list of candidate matches computed by VPR as input. The LPR module aims to further validate the candidate match set by computing the similarity of the surrounding point clouds to the candidate matches using a lidar descriptor. This work proposes the use of the ScanContext~\cite{scancontext} descriptor, however, we emphasize that any other lidar place recognition method can be used instead given the modular design of the pipeline\cite{bow3d, pointnetvlad}.

The first step of the LPR module is to retrieve the point cloud with the closest timestamp for each candidate match. Due to the VLP16's relative sparsity in single scans (only 16 vertical beams), the proposed algorithm retrieves the aggregated cloud around the reference cloud (keyframe). An aggregated cloud contains $2R + 1$ individual clouds that get transformed into the keyframes reference frame using the map's trajectory. Assume we have a timestamp from a candidate match pair $t_1^i$, we then find the closest point cloud from $M_R$, $S_{M_R}^{i'}$, where $i'$ is the closest timestamp to $i$, as well as the $R$ neighbouring clouds $S_{M_R}^{i'\pm r}$. To transform each point cloud into the keyframes reference frame we use equation \ref{eq:4.4}.

\begin{equation} \label{eq:4.4}
(S_{M_R}^{i'\pm r})\prime = (T_{L_{i'}}^{W_R})^{-1} \, T_{L_{i'\pm r}}^{W_R} \, S_{M_R}^{i'\pm r}
\end{equation}

After each cloud is transformed into the middle cloud frame, they are all aggregated into a single point cloud. This process is repeated for both timestamps in a candidate match. The next step is to compute the ScanContext descriptor for each aggregated scan. Two distances can be computed from ScanContext descriptors, the first distance is computed by summing the distances between the corresponding columns, and the second distance calculates the minimum of the first distance of every possible column shift in the descriptor. Since the ScanContext descriptor is not viewpoint invariant (i.e. two scans of the same area can give very different scores based on their viewpoint), we use the slower but more robust second distance metric. If a candidate matches ScanContext distance is below a set threshold, $\psi$, then the candidate match is accepted. This module then outputs a filtered list of matches in the same format as equation \ref{eq:4.3}.

\subsection{Map-to-Map Scan Registration} \label{sec:map-registration}

After the candidate matches have been filtered with LPR, the next step is to estimate the relative poses between the maps at the candidate match timestamps. We first obtain a coarse estimate of the transform $T_{W_2}^{W_1}$ by using the transforms for each lidar scan in their world frames, $T_{L_i}^{W_1}$ and $T_{L_j}^{W_2}$ using equation \ref{eq:4.5}. This initial estimate is a coarse estimate since the matches provided by VPR and LPR are not necessarily at the same location.

\begin{equation} \label{eq:4.5}
\widehat{T_{W_T}^{W_R}} = T_{L_i}^{W_R} * (T_{L_j}^{W_T})^{-1}
\end{equation}

Then the same process outlined in \ref{sec:lpr} retrieves an aggregate cloud around each timestamp. However, each cloud is transformed into their respective world frames $W^R$ and $W^T$. Scan registration is then performed to refine the initial coarse estimate given in \ref{eq:4.5}, resulting in an updated transform $T'$, using these aggregate scans. The final estimate of $T_{W_T}^{W_R}$ is computed using equation \ref{eq:4.6}.

\begin{equation} \label{eq:4.6}
T_{W_T}^{W_R} = T' * \widehat{T_{W_T}^{W_R}} 
\end{equation}

The scan registration algorithm used is known as Generalized-ICP (GICP) \cite{gicp}, which is an extension of the original ICP algorithm \cite{icp}. However, GICP incorporates an estimate of the sensor noise to consider the uncertainty in each point measurement into the optimization, making it more robust to noise. Additionally, GICP introduces the use of plane-to-plane error, taking into account the geometric structure of both scans when optimizing for the transform between them. GICP is shown to be more robust to outliers than regular ICP, making it very suitable for this application\cite{gicp}. If the resulting fitness score from the scan registration is larger than a threshold $\xi$, the match is discarded. For each match, the best registration result is retained. The goal of this step is to store an estimate of the relative transform between maps for every candidate match, thus computing a trajectory of relative transforms where each timestamp in $M_T$ has an associated transform $T_{W_T}^{W_R}$. 

\subsection{Non-Rigid Map Alignment}

\begin{figure}[h]
    \centering
    \vspace{5pt}
    \includegraphics[width=0.45\textwidth]{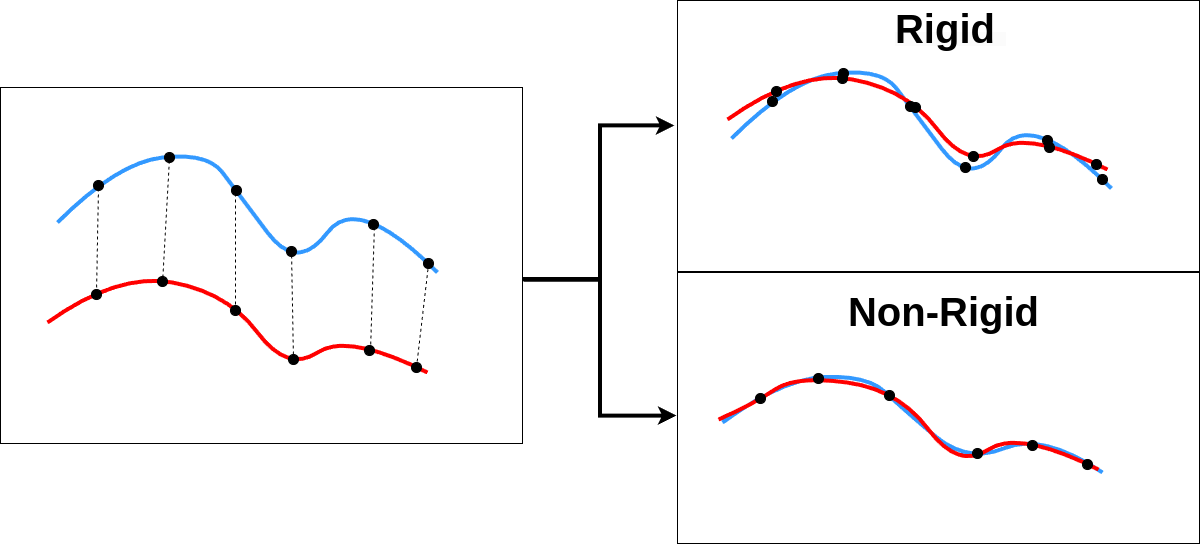}
    \caption{Starting map pair with correspondences (left), rigid map alignment (top right) and non-rigid map alignment (bottom right).}
    \label{fig:nonrigidalignment}
\end{figure}

The final step of the map alignment process is to apply the relative trajectory estimate computed in \ref{sec:map-registration} to the trajectory in $M_T$ to convert to the world frame of $M_R$. However, SLAM estimates, even with loop closures, can still drift over time; therefore, the relative transform between the maps can differ depending on where in the maps an association is found, resulting in a non-constant relative trajectory. Knowing this, we aim to apply the relative trajectory adaptively over $M_T$'s trajectory, resulting in a non-rigid alignment. This is the primary reason for scan registration not being performed on the full point cloud from each map, as this process will produce a rigid alignment between maps, resulting in misalignment. This difference is illustrated in figure \ref{fig:nonrigidalignment}. 

To perform the non-rigid alignment, we determine the associated map-to-map transform, $T_{W_T}^{W_R}$, for every pose in $M_T$'s trajectory by interpolating from the relative trajectory. A B-Spline\cite{spline} is estimated for the relative trajectory to interpolate over the trajectory. For each pose in the original trajectory, the interpolated relative pose is applied to the original pose in $M_T$'s trajectory to transform it into the $W_R$ frame. The interpolation also has the added benefit of smoothing the resulting trajectory and reducing the effect of noise in the estimates.

\section{RESULTS}
\label{section:results}
\subsection{Experimental Data}

\begin{figure}[!h]
    \centering
    \vspace{5pt}
    \includegraphics[width=0.45\textwidth]{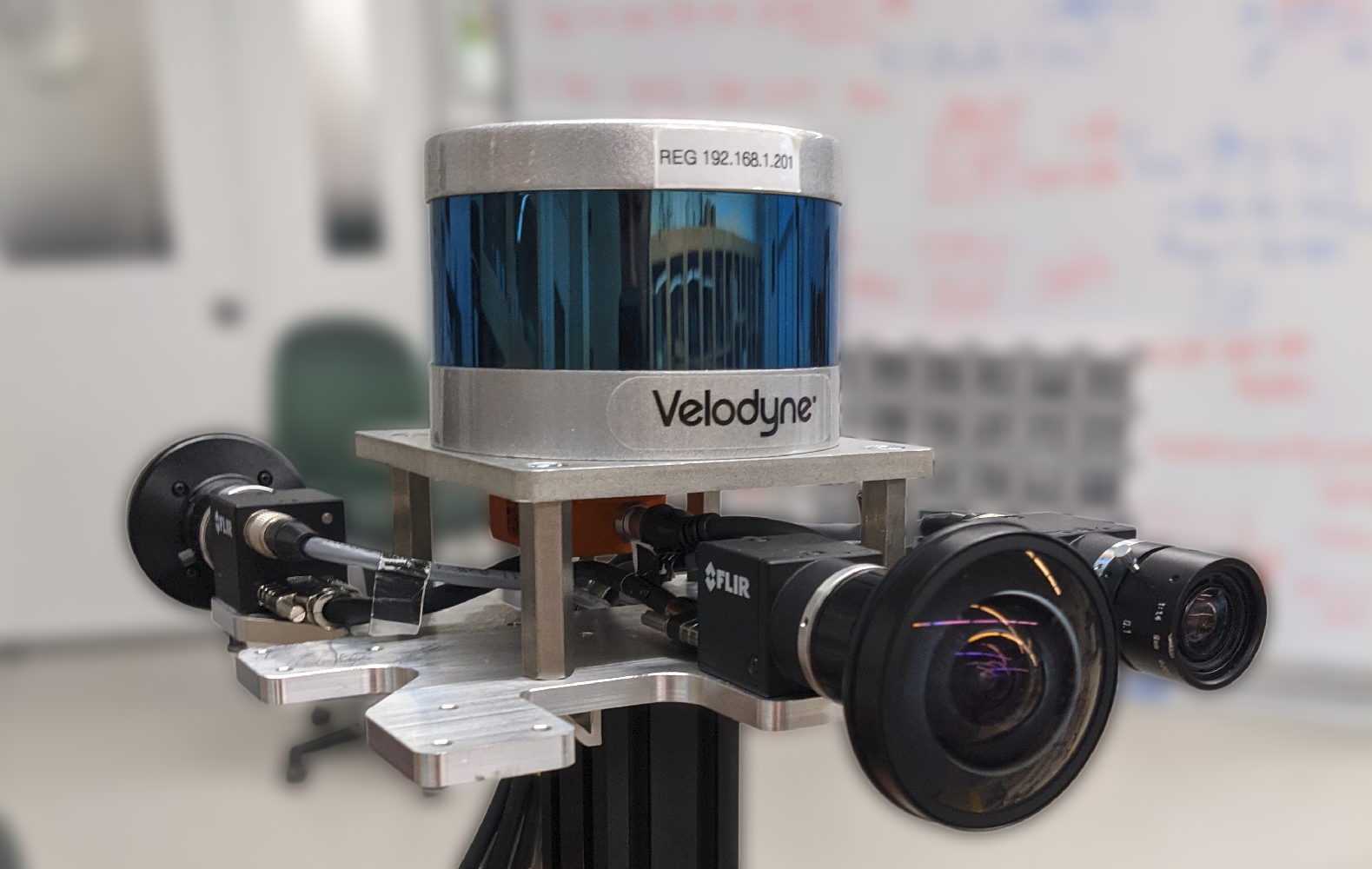}
    \caption{Data collection device outfitted with Velodyne lidar, 3 Flir cameras and Xsens IMU.}
    \label{fig:ig_handle}
\end{figure}

The data required to implement the proposed map alignment strategy is time-synchronized visual and LiDAR data, as well as an initial estimate of a sensor trajectory. For data collection, a modular platform shown in figure \ref{fig:ig_handle} was developed, capable of being carried by a single inspector or being mounted on various types of robotic platforms. The platform is equipped with two Flir Blackfly USB3 RGB cameras running at 20hz (one with a fisheye lens and one with a low FOV lens), an Xsens MTi-30 AHRS IMU running at 200hz and a Velodyne VLP16 lidar running at 10hz. 

In the experimental evaluation, two distinct datasets gathered using this device were employed. The first dataset was gathered from the Structures Laboratory at the University of Waterloo, spanning three unique acquisition instances. Each acquisition was spaced with varying temporal intervals to test the robustness of our approach across different time spans. This dataset is an example of an indoor environment, rich in visual and lidar data. The second dataset was gathered from the Conestogo Bridge in Kitchener, Ontario. Comprising of three instances, this dataset serves as an outdoor example, markedly sparse in distinct visual and lidar features, making it a more challenging test case for validation. To collect 3D maps and trajectories of each area, SC-A-LOAM (a combination of \cite{loam} as odometry and \cite{scancontext} for loop closure) was used. The details of each dataset are outlined in table \ref{table:datasets}, where $SL_i$ represents a collection from Structures Lab dataset and $CB_i$ from the Conestogo Bridge dataset.

\begin{table}[h]
\caption{Details on data in each individual mapping session. $SL_i$ refers the the i'th session from the Structure Lab dataset, and $CB_i$ is from the Conestogo Bridge dataset.}
\label{table:datasets}
\begin{center}
    \begin{tabular}{ccccc}
    \hline
    Dataset  & Date   & Duration   & Scans      &  Images  \\ \hline
    $SL_1$    & Sep 30 2021 & 78s  & 775 & 152 \\ 
    $SL_2$    & Oct 21 2021 & 75s  & 752 & 1479 \\ 
    $SL_3$    & Oct 27 2021 & 97s  & 967 & 1909  \\ 
    $CB_1$    & Oct 27 2021 & 228s & 2285 & 4564  \\ 
    $CB_2$    & Oct 27 2021 & 228s & 2266 & 4533 \\ 
    $CB_3$    & Oct 27 2021 & 182s & 1813 & 3615  \\ \hline
    \end{tabular}
\end{center}
\end{table}

\subsection{Experimental Results}
\label{subsection:experimental_results}

The goal of map alignment, in the context of repeated infrastructure inspections, is to maximize the overlap between two 3D maps. Therefore, to assess a map alignment approach we use three metrics. The first is point-to-point error between maps, which is used in the ICP \cite{icp} algorithm to optimize for the transform between clouds. This error can be unreliable at times since it is greedy in terms of its point association (i.e. it uses the nearest point in cloud 2 to the reference point in cloud 1 to compute error). It is useful for determining obviously bad alignments but sometimes performs poorly at comparing between good alignments, especially if the maps span different areas. The last two metrics are CloudCompare's \cite{cloud_compare} surface and volume density metrics. The surface density divides the number of neighbours ($N$) for each point (within a radius) by the neighbourhood surface ($\pi R^2$), while the volume density divides by the neighbourhood volume (${3 \pi R^3}/4$). This results in a density score for each point in the cloud. We use a density metric as a measure for map alignment because poorly aligned maps have more noise than well aligned ones.

\begin{figure}[h]
\centering
\begin{subfigure}{.24\textwidth}
  \centering
  \vspace{10pt}
  \includegraphics[width=0.95\linewidth]{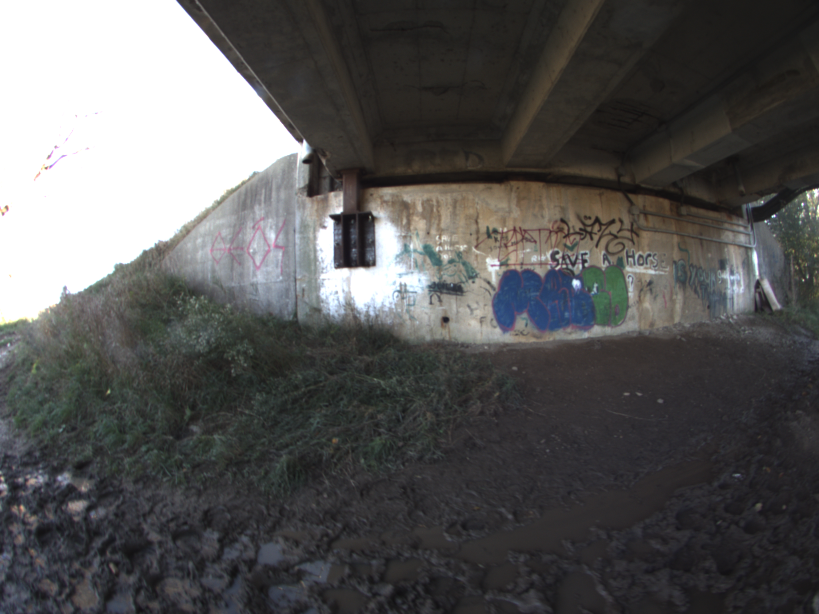}
\end{subfigure}%
\begin{subfigure}{.24\textwidth}
  \centering
  \vspace{10pt}
  \includegraphics[width=0.95\linewidth]{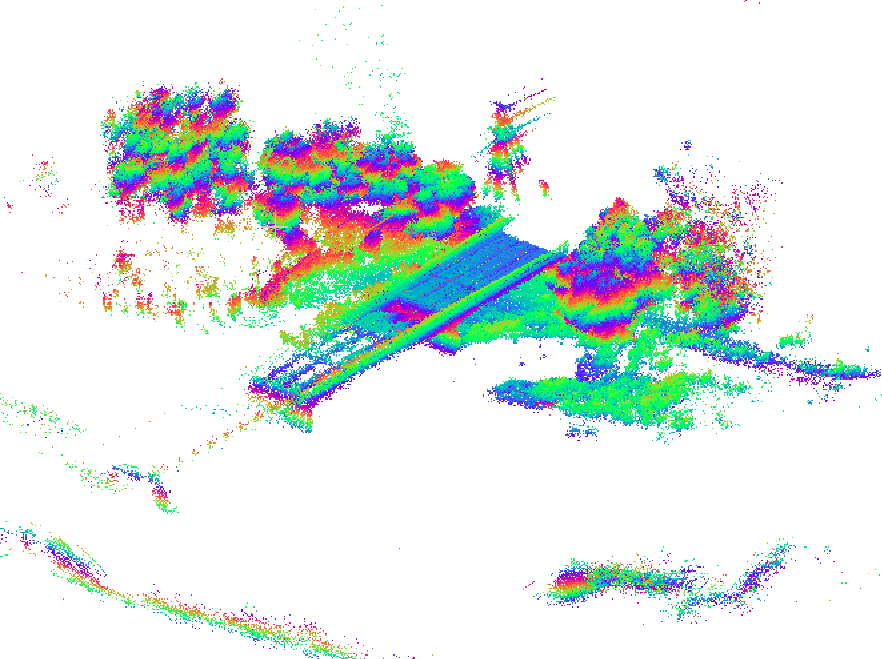}
\end{subfigure}
\caption{Image from Conestogo Bridge dataset (left) along with the 3D map (right).}
\label{fig:datasets}
\end{figure}

\begin{figure}[h]
\centering
\begin{subfigure}{.24\textwidth}
  \centering
  \includegraphics[width=0.95\linewidth]{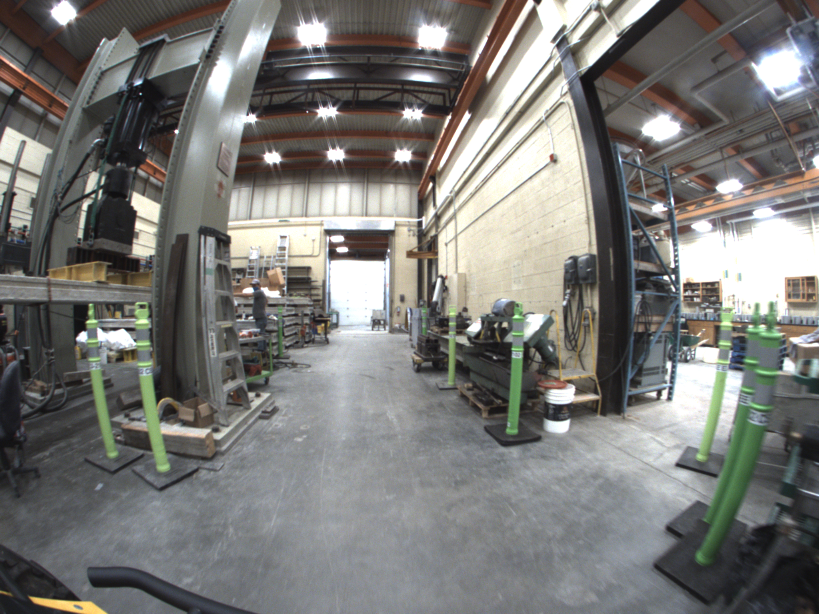}
\end{subfigure}%
\begin{subfigure}{.24\textwidth}
  \centering
  \includegraphics[width=0.95\linewidth]{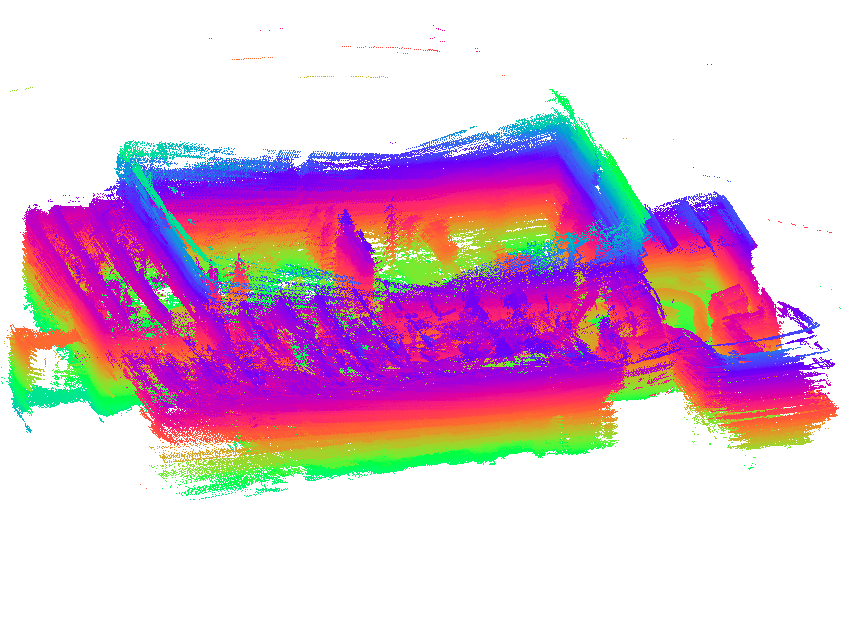}
\end{subfigure}
\caption{Image from Structures Lab dataset (left) along with the 3D map (right).}
\label{fig:datasets}
\end{figure}
 
 The following experiments test pairs of datasets, $DS_i$-$DS_j$, where $DS_i$ is the target map to and $DS_j$ is the reference map. For the SL dataset, we empirically choose $SL_2$ as the reference map since it covers the most visual area. For the CB dataset, $CB_1$ is chosen for the same reason. We compare 3 methods of map alignment: 1) GICP \cite{gicp}, 2) VMA (Visual Map Alignment), 3) LMA (Lidar Map Alignment) and 4) VLMA (Visual-Lidar Map Alignment). GICP serves as a reference for a naive approach to map alignment, which would be to attempt scan registration between entire maps. VMA uses DBoW\cite{dbow2} to find the best image match between maps, which is used to initialize scan registration and solve for a single transform between maps. LMA uses ScanContext \cite{scancontext} to find the best possible lidar scan match between maps, and uses it to align the maps similarly to VMA. Lastly, VLMA is the approach that has been introduced in \ref{section:meth}.

 We cannot compare directly against any existing MSS or localization packages due to the requirement of running their specific SLAM (or other map-building) method. This is not always possible in the context of long-term infrastructure inspections, as previous data may have had different sensors which the SLAM method does not support, or the environment itself may not be suitable for their algorithm. Additionally, data may be compressed or sub-sampled to reduce storage usage, and this subset of the data is often not sufficient for most SLAM methods.

The first two tests compute the average surface density (ASD) and average volume density (AVD) of the alignment using a 0.2m radius. As mentioned previously, the surface density metric computes a score for each point in the cloud; here, we take an average of all the surface densities to score the alignment result.

\begin{table}[h]
% \vspace{15pt}
\caption{Comparison of the average surface density (ASD) of each map alignment method, using a 0.2m radius.}
\label{table:surface_density}
\begin{center}
    \begin{tabular}{ccccc}
    \hline
    Map Pair   &  GICP     & VMA      & LMA   & VLMA (ours)  \\ \hline
    $SL_2$-$SL_1$   & 1198.29	&1196.3	&1198.23	&\bf{1212.37} \\
    $SL_2$-$SL_3$   & 1237.49	&1224.18	&1113.65	&\bf{1229.19} \\ 
    $CB_1$-$CB_2$   & 1054.21	&1174.22	&1121.81	&\bf{1185.65} \\ 
    $CB_1$-$CB_3$   & 1055.06	&1122.08	&1064.56	&\bf{1108.64}\\ \hline
    \end{tabular}
\end{center}
\end{table}

\begin{table}[h]
\vspace{10pt}
\caption{Comparison of the average volume density (AVD) of each map alignment method, using a 0.2m radius.}
\label{table:volume_density}
\begin{center}
    \begin{tabular}{ccccc}
    \hline
    Map Pair   &  GICP     & VMA      & LMA   & VLMA (ours)  \\ \hline
    $SL_2$-$SL_1$   & 4493.6	&4486.11	&4493.37	&\bf{4546.39} \\ 
    $SL_2$-$SL_3$   & 4493.61	&4590.69	&4176.19	&\bf{4609.45} \\ 
    $CB_1$-$CB_2$   & 3953.29	&4403.31	&4206.78	&\bf{4446.2} \\ 
    $CB_1$-$CB_3$   & 3956.47	&4207.8	    &3992.09	&\bf{4157.41}\\ \hline
    \end{tabular}
\end{center}
\end{table}

% comment on tests + introduce the test below
In table \ref{table:surface_density} and table \ref{table:volume_density} we see that VLMA performs best (highest density), with VMA performing second best, followed by GICP and LMA performing similarly. This shows that lidar alone does not perform the map alignment task well, however it adds crucial information that visual sensors lack, further increasing robustness and improving map alignment accuracy of visual map alignment. 

\begin{table}[h]
\caption{Point-to-Point error of different map alignment methods.}
\label{table:point_error}
\begin{center}
    \begin{tabular}{ccccc}
    \hline
    Map Pair  &  GICP     & VMA      & LMA   & VLMA (ours)  \\ \hline
    $SL_2$-$SL_1$   & 0.7514	&0.1760	&0.1449	&\bf{0.1221}\\ 
    $SL_2$-$SL_3$   & 0.1292	&0.1247	&6.5424	&\bf{0.1026} \\ 
    $CB_1$-$CB_2$   & 1.363	    &0.1248	&0.1243	&\bf{0.1217} \\ 
    $CB_1$-$CB_3$   & 1.6262	&0.2473	&1.6721	&\bf{0.2169}\\ \hline
    \end{tabular}
\end{center}
\end{table}

In table \ref{table:point_error}, we see that GICP consistently performs poorly due to convergence issues (to the correct transform), especially if the initial offset between maps is large. LMA also under-performs in two datasets ($SL_2$-$SL_3$ and $CB_1$-$CB_3$), showing its inconsistency in finding a true positive match between maps. VMA performs well in every test. However, it is still outperformed by our algorithm, VLMA, in every test, showing that utilizing lidar data helps ensure more robust and accurate visual map alignment.

\begin{figure}[h]
\centering
\begin{subfigure}{.24\textwidth}
  \centering
  \includegraphics[width=0.95\linewidth]{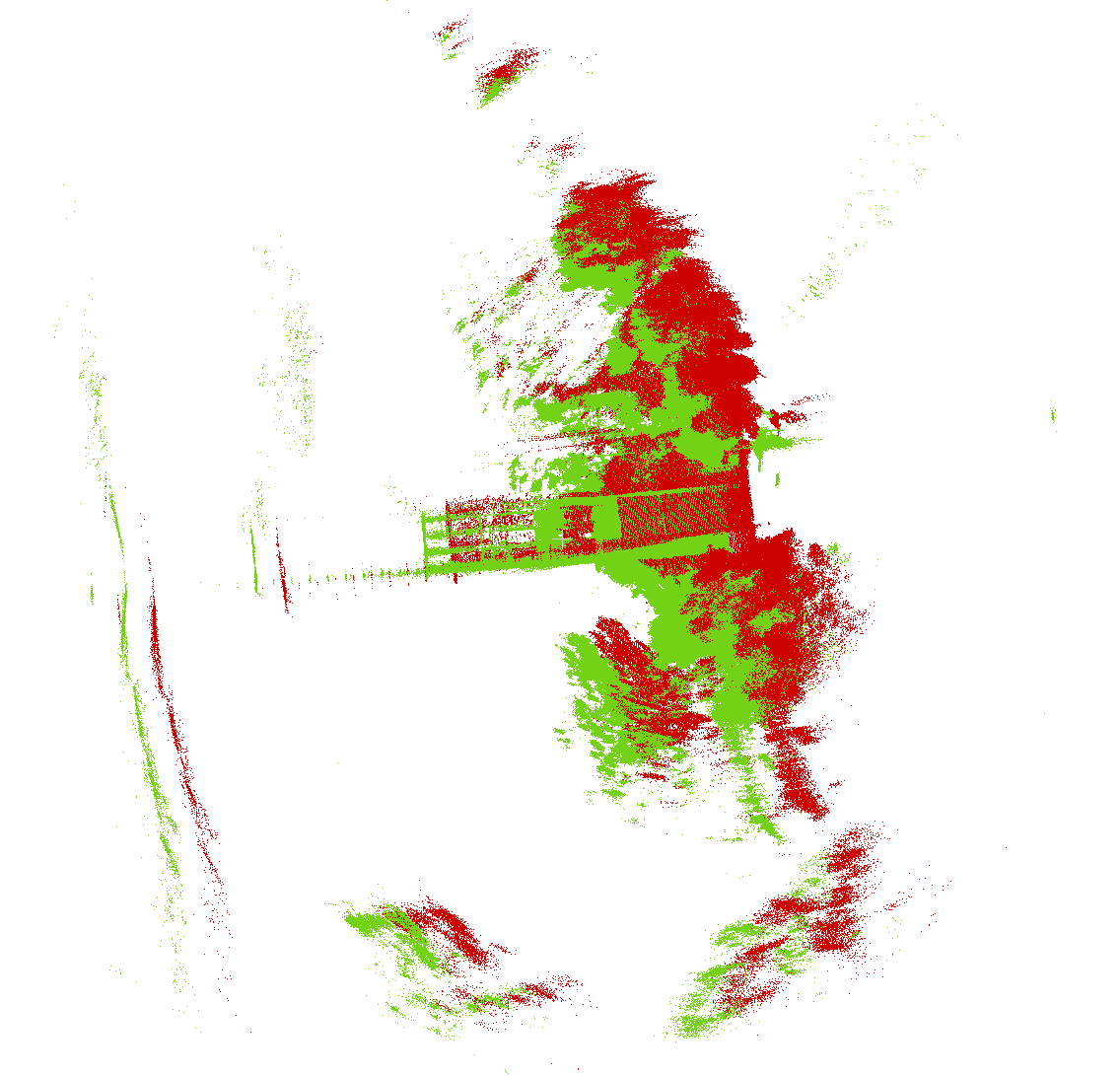}
\end{subfigure}%
\begin{subfigure}{.24\textwidth}
  \centering
  \includegraphics[width=0.95\linewidth]{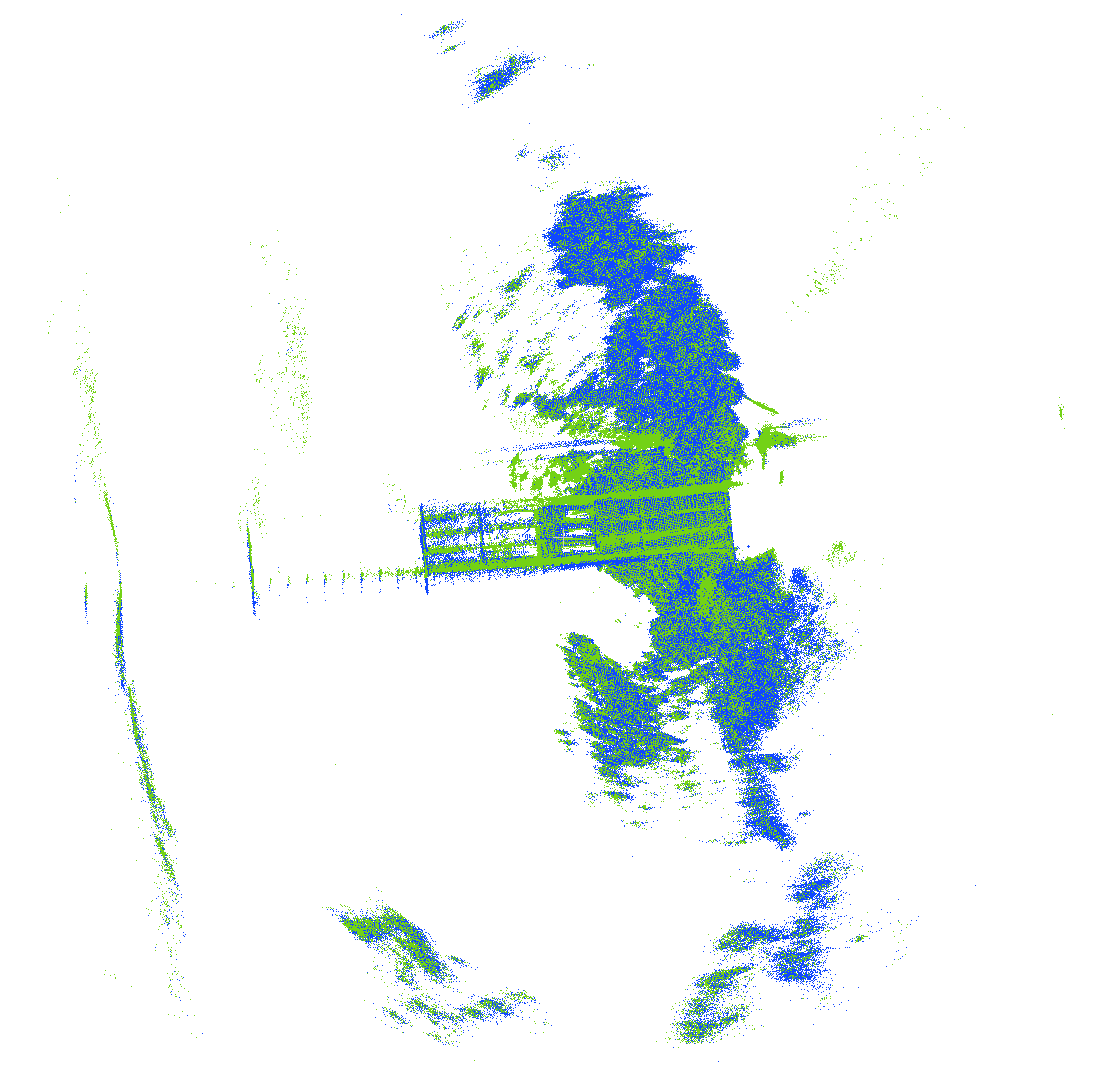}
\end{subfigure}
\caption{Comparison between map alignment methods, aligning to the reference cloud (green), GICP (left) and ours (right).}
\label{fig:datasets}
\end{figure}

\subsection{Ablation Study}

To study the effects of the two methods of outlier detection that the algorithm uses: ScanContext (SC) filtering and geometric outlier verification (Geo.), we perform an ablation study. The ablation study runs the same map alignment tests that were shown in \ref{subsection:experimental_results}, however using only VLMA, and with each possible combination of outlier methods employed. The metric gathered was the mean of the ASD's (MASD) and AVD's (MAVD) from each test.

\begin{table}[h]
\caption{Outlier rejection ablation study, comparing the  mean of all alignments ASD (MASD) and AVD (MAVD) with and without the different outlier rejeciton methods.}
\label{table:outlier_ablation}
\begin{center}
    \begin{tabular}{ccccc}
    \hline  \multicolumn{3}{c}{ Outlier Detection } & \multicolumn{2}{c}{ Metric } \\
    & SC & Geo. & MASD & MAVD  \\
    \hline  
    & - & - & 1181.27 & 4429.77  \\
    & $\checkmark$ & - & 1186.26 & 4448.47 \\
    & - & $\checkmark$ & 1181.08 & 4429.05  \\
    & $\checkmark$ & $\checkmark$ & $\mathbf{1188.28}$ & $\mathbf{4456.03}$ \\
    \hline
    \end{tabular}
\end{center}
\end{table}

Table \ref{table:outlier_ablation} shows that using just SC outperforms using just Geo., however using them together results in a better density than using either alone, again validating the proposed algorithm.

\begin{table}[h]
\caption{Rigid vs non-rigid alignment in 3 metrics: average surface density (ASD), average volume density (AVD) and point-to-point error (P2P).}
\label{table:rigid_v_nonrigid}
\begin{center}
    \begin{tabular}{ccccc}
    \hline Method & Map Pair & ASD & AVD & P2P \\ 
    \hline \multirow{4}{*}{ Rigid } & $SL_2$-$SL_1$ & 1204.96 & 4518.61 & 0.1487 \\
    & $SL_2$-$SL_3$ & 1203.76 & 4514.11& 0.2228 \\
    & $CB_1$-$CB_2$ & 1211.2 & 4542.01 & 0.1366 \\
    & $CB_1$-$CB_3$ & 1121.65 & 4206.18 & 0.2470 \\
    \hline \multirow{4}{*}{ Non-Rigid } & $SL_2$-$SL_1$ & 1210.47 & 4546.39 & 0.1385 \\
    & $SL_2$-$SL_3$ & 1225.5 & 4609.45 & 0.1026 \\
    & $CB_1$-$CB_2$ & 1186.9 & 4462.32 & 0.1217 \\
    & $CB_1$-$CB_3$ & 1122.16 & 4205.97 & 0.2469 \\
    \hline
    \end{tabular}
\end{center}
\end{table}

\begin{figure}[!h]
    \centering
    \includegraphics[width=0.45\textwidth]{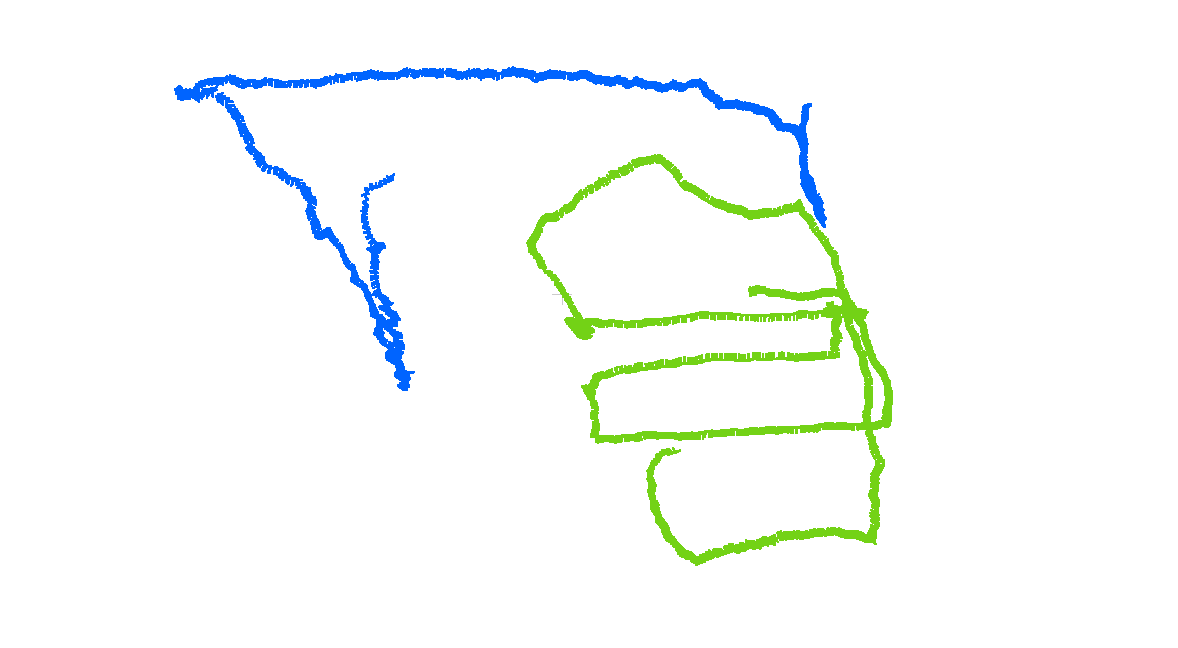}
    \caption{Top down view of $CB_1$ trajectory (green) and aligned $CB_3$ trajectory (blue).}
    \label{fig:traj_comaprison}
\end{figure}

The last experiment compares using rigid alignment versus non-rigid alignment on the same datasets. Table \ref{table:rigid_v_nonrigid} shows that non-rigid alignment outperforms rigid alignment in every dataset except for $CB_1$-$CB_3$, where it is nearly a tie. Investigating this dataset further shows that there was only one instance of map overlap, as displayed in figure \ref{fig:traj_comaprison}, that was sufficiently similar to estimate a relative transform between maps. Since only one instance exists, then the resulting alignment was relatively static over the whole map, making it roughly equivalent to using rigid alignment.

\section{CONCLUSIONS}
\label{section:conclusion}

This paper addresses the gap in multi-session mapping for infrastructure inspection automation by introducing a map alignment algorithm capable of performing independently of SLAM and using the combination of visual and lidar data for improved map alignment robustness. Additionally, we show that the non-rigid map alignment formulation improves map overlap, which is crucial for tracking infrastructure health over time. We validate this with experimental data on two datasets in different environments, including scans inside a dynamic structural laboratory, and scans of a concrete bridge. This new approach to map alignment facilitates the use of cutting-edge SLAM methods while retaining the ability to keep multiple maps spatially aligned over time. The code for this project is available for download at https://github.com/jakemclaughlin6/vlma. 

% This flexibility facilitates the use of cutting-edge SLAM methods, regardless of their inherent multi-session capabilities. Our method outperforms the GICP approach in point cloud map alignment accuracy. Sole reliance on visual place recognition weakened the method's performance. However, integrating visual geometric outlier detection or lidar outlier detection using lidar descriptors significantly improved it, with the best outcomes achieved using both. By demonstrating that map alignment can be done offline, the dependency on specific SLAM algorithms and systems is reduced. This research integrates both visual and lidar-based recognition, enhancing map alignment accuracy. However, the method's limitations include manual parameter tuning, its offline nature, and reliance on accurate initial maps. Moreover, its robustness is yet to be validated across diverse scenarios. Future work should consider integrating this method into online SLAM systems, delve into advanced recognition techniques, and explore static object associations. In essence, while our study advances map alignment, highlighting its potential, there remains ample scope for refinement and exploration in the field.

% \addtolength{\textheight}{-12cm}   
% This command serves to balance the column lengths on the last page of the document manually. It shortens the textheight of the last page by a suitable amount. This command does not take effect until the next page so it should come on the page before the last. Make sure that you do not shorten the textheight too much.

\label{section:references}
\bibliography{cited_works}
\bibliographystyle{ieeetr}

\end{document}